\newif\ifijcai
\ijcaifalse

\ifijcai
  \documentclass{article}
  \usepackage{ijcai19}
  \usepackage[hidelinks]{hyperref}
\else
 \documentclass[11pt, a4paper, logo]{deepmind}
\fi

\usepackage{todonotes}
\usepackage{natbib}

\usepackage{subcaption}
\usepackage{times}
\usepackage{soul}
\usepackage{url}
\usepackage[capitalise,noabbrev]{cleveref}
\usepackage[utf8]{inputenc}
\usepackage{graphicx}
\usepackage{amsmath}
\usepackage{booktabs}
\usepackage{float}
\usepackage[symbol]{footmisc}
\usepackage{tikz}
\usetikzlibrary{bayesnet,fit,positioning,decorations.shapes}
\tikzset{
utility/.style = {fill=yellow!50,diamond, rounded corners=0},
trueutility/.style = {fill=green!50,diamond, rounded corners=0},
decision/.style = {fill=blue!10,rectangle,rounded corners=0},
information/.style = {dotted},
control incentive/.style = {red, dashed, thick},
information incentive/.style = {ForestGreen, dashdotted, thick},
observation incentive/.style = {blue, dotted, ultra thick},
downplay/.style = {color=gray,thin},
highlight/.style = {red, thick},
surprise/.style = {dashed},
counterfactual/.append style = {dashed},
player1/.style = {fill=red!20}, 
player1/.default = {-20},
player2/.style = {fill=green!20}, 
player2/.default = {-20},
player3/.style = {fill=teal!20}, 
player3/.default = {-20},
player4/.style = {fill=black!20}, 
player4/.default = {-20},
common/.style = {fill=brown!20},
}

\newcommand{\rf}{\Theta^{\mathrm{R}}}
\newcommand{\tf}{\Theta^{\mathrm{T}}}
\newcommand{\hf}{\Theta^{\mathrm{H}}}
\definecolor{darkblue}{HTML}{191970}

\title{Modeling AGI Safety Frameworks with Causal Influence Diagrams}

\ifijcai

\author{%
Tom Everitt$^\dagger$\footnote{Contact author, tomeveritt@google.com}\and 
Ramana Kumar$^\dagger$\and 
Victoria Krakovna\footnote{equal contribution}\and 
Shane Legg
\affiliations
DeepMind
}
\date{}

\begin{document}

\maketitle

\else

\author[1]{Tom Everitt$^*$}
\author[1]{Ramana Kumar$^*$}
\author[1]{Victoria Krakovna\footnote{equal contribution}}
\author[1]{Shane Legg}

\affil[1]{DeepMind}
\paperurl{}

\fi

\begin{abstract}
    Proposals for safe AGI systems are typically made at the level of \emph{frameworks}, specifying how the components of the proposed system should be trained and interact with each other.
    In this paper, we model and compare the most promising AGI safety frameworks using \emph{causal influence diagrams}.
    The diagrams show the optimization objective and causal assumptions of the framework.
    The unified representation permits easy comparison of frameworks and their assumptions.
    We hope that the diagrams will serve as an accessible and visual introduction to the main AGI safety frameworks.
\end{abstract}

\ifijcai\else
  \begin{document}
  \maketitle
  \balance
\fi

\section{Introduction}

One of the primary goals of AI research is the development of artificial agents that can exceed human performance on a wide range of cognitive tasks, in other words, artificial general intelligence (AGI).
Although the development of AGI has many potential benefits, there are also many safety concerns that have been raised in the literature~\citep{bostrom2014superintelligence,Everitt2018litrev,Amodei2016concrete}.
Various approaches for addressing AGI safety have been proposed~\citep{Leike2018alignment, Christiano2018ida, Irving2018debate, Hadfield-Menell2016cirl, Everitt2018thesis}, often presented as a modification of the reinforcement learning (RL) framework, or a new framework altogether.

Understanding and comparing different frameworks for AGI safety can be difficult because they build on differing concepts and assumptions.
For example, both reward modeling \citep{Leike2018alignment} and cooperative inverse RL \citep{Hadfield-Menell2016cirl} are frameworks for making an agent learn the preferences of a human user, but what are the key differences between them?
In this paper, we show that \emph{causal influence diagrams} (CIDs) \citep{Koller2003,Howard1984} offer a unified way of clearly representing the most prominent frameworks discussed in the AGI safety literature.
In addition to surfacing many background assumptions, the unified representation enables easy comparison of different frameworks.

What makes this unified representation possible is the high information density of CIDs.
A CID describes safety frameworks with random variables that together specify what is under an agent's control, what the agent is trying to achieve by selecting outcomes under its control, and what information the agent has available when making these decisions.
Multi-agent CIDs can model the interaction of multiple agents.
CIDs also encode the causal structure of the environment, specifying which variables causally influence which others.
This causal information reveals much about agents' abilities and incentives to influence their environment \citep{Everitt2019incentives1}.

In our terminology, a \emph{framework} is a way of building or training an AI system, or a way of situating the system in an environment so that it learns or achieves some goal.
We take a high-level view on what counts as a framework, so we can talk about different approaches to building safe AGI.
Frameworks determine what agents know and want, insofar as we can take the intentional stance towards them.
We focus on frameworks rather than on implementations of specific algorithms in order to generalize across different safety problems and draw broad lessons. 
A \emph{safety framework} is a framework designed to highlight or address some problem in AGI safety.

The actual behaviour of artificial agents will be contingent on implementation details such as the training algorithms used or the task specification.
But these details can obscure general patterns of behaviour shown by intelligent systems within a particular framework.
By focusing on frameworks, such as reinforcement learning (RL) or RL with reward modeling, we can characterize the behaviour to be expected from increasingly intelligent agents (and highlight any safety concerns).
Modeling frameworks with CIDs enables analysis of high-level concerns such as agent incentives
before getting into the details of training within a framework.

We describe the framework models in \cref{sec:mdp-based,sec:qa-systems}.
The descriptions of the frameworks are necessarily brief; we refer the reader to the original papers for more detailed descriptions.
In \cref{sec:discussion} we discuss modeling choices and interpretations.

\section{MDP-Based Frameworks}
\label{sec:mdp-based}

In this section, we look at frameworks based on Markov Decision Processes (MDPs), which covers several of the safety frameworks in the literature.
Afterwards (\cref{sec:qa-systems}), we look at frameworks based on question-answering systems, covering most of the other frameworks so far proposed.

\subsection{RL in an MDP}

Our basic example of a framework is standard reinforcement learning (RL) in a Markov Decision Process (MDP) \citep{Sutton2018}.
This framework is not intended to address any particular safety concerns.
It is, however, the foundation for most present-day development of artificial agents, and will serve as a familiar skeleton on which many of our models of other frameworks will be based.

\begin{figure}[ht]
    \centering
    \begin{tikzpicture}[
        node distance=0.7cm,
        every node/.style={
            draw, circle, minimum size=0.8cm, inner sep=0.5mm}]
        \node (R1) [utility] {$R_1$};
        \node (S1) [above = of R1] {$S_1$};
        \node (A1) [right = of R1, decision] {$A_1$};
        \node (R2) [right = of A1, utility] {$R_2$};
        \node (S2) [above = of R2] {$S_2$};
        \node (A2) [right = of R2, decision] {$A_2$};
        \node (R3) [right = of A2, utility] {$R_3$};
        \node (S3) [above = of R3] {$S_3$};
        
        \edge {S1} {R1};
        \edge {S2} {R2};
        \edge {S3} {R3};
        \edge {A1,S1} {S2};
        \edge {A2,S2} {S3};
        \edge[information] {S1} {A1};
        \edge[information] {S2} {A2};
    \end{tikzpicture}
   \caption{RL in an MDP}
   \label{fig:rl-mdp}
\end{figure}
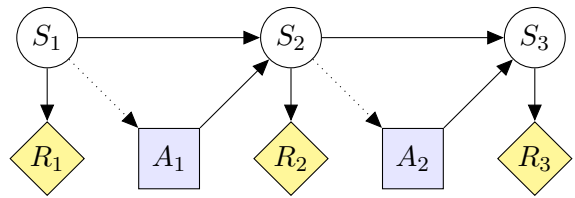

\Cref{fig:rl-mdp} models an MDP with a CID.
A CID is a causal graph or Bayesian network \citep{Pearl2009} with special decision and utility nodes.
Decision nodes are drawn as squares, and utility nodes as diamonds.
Edges represent causal influence, except for those going into decision nodes.
The latter are called \emph{information links}, and drawn with dotted lines.
The agent's decision for a particular decision node $A$ can only be based on the parents of the decision nodes, i.e., on the nodes that have information links to $A$.
For example, in \cref{fig:rl-mdp}, the choice of an action can only depend on the current state.

For this foundational example, we explain the random variables in detail.
The MDP modelled in \cref{fig:rl-mdp} has transition function $T(s,a)$ giving a distribution over next states after taking action $a$ in state $s$, and reward function $R(s)$ giving the reward obtained for entering state $s$.
The random variables in the model are:
\begin{itemize}
    \item $S_i$: the environment state at timestep $i$. $S_1$ is drawn from the initial state distribution, and $S_{i+1}$ is drawn according to $T(S_i, A_i)$.
    \item $A_i$: the agent action at timestep $i$.
    \item $R_i$: the reward for timestep $i$, drawn from $R(S_i)$.
\end{itemize}

The CID has a repeating pattern in the relationships between the variables, if we abstract over the timestep index $i$.
We show $3$ timesteps to make the pattern obvious, but this is an arbitrary choice: we intend to represent trajectories over any number of timesteps.
Formally, this can be done by extending the diagram with more variables in the obvious way.

\Cref{fig:rl-mdp} represents a \emph{known} MDP, where the transition and reward probability functions ($T$ and $R$) are known to the agent.
In this case, an optimal policy only needs access to the current state, as the Markov property of the MDP makes the past irrelevant.
For this reason, we have chosen \emph{not} to include information links from past states and rewards, though adding them would also constitute a fair representation of an MDP.

\begin{figure}[ht]
  \centering
  \begin{tikzpicture}[node distance=0.6cm]
      \tikzset{
        every node/.style={ draw, circle, minimum size=0.8cm, inner sep=0.5mm }
      }
      \node (R1) [utility] {$R_1$};
      \node (S1) [above = of R1] {$S_1$};
      \node (A1) [right = of R1, decision] {$A_1$};

      \node (R2) [right = of A1, utility] {$R_2$};
      \node (S2) [above = of R2] {$S_2$};
      \node (A2) [right = of R2, decision] {$A_2$};

      \node (R3) [right = of A2, utility] {$R_3$};
      \node (S3) [above = of R3] {$S_3$};

      \path
      (S1) edge[->] (R1)
      (S1) edge[->, information] (A1)
      (R1) edge[->, information] (A1)

      (S1) edge[->] (S2)
      (A1) edge[->] (S2)
      (A1) edge[->, information, bend right] (A2)
      (R1) edge[->, information, bend right] (A2)

      (S2) edge[->] (R2)
      (S2) edge[->, information] (A2)
      (R2) edge[->, information] (A2)

      (S2) edge[->] (S3)
      (A2) edge[->] (S3)

      (S3) edge[->] (R3)
      ;
      
      \node (ah) [minimum size=0mm,node distance=2mm, below left = of R1, draw=none] {};

      \draw[information]
      (S1) edge[ in=135,out=-120] (ah.center)
      (ah.center) edge[->, out=-45,in=-150] (A2);

      \node (theta) [above = of S1] {$\tf$};
      \path
      (theta) edge[->] (S1)
      (theta) edge[->] (S2)
      (theta) edge[->] (S3)
      ;
      \node (R) [below = of R1] {$\rf$};
      \edge  {R} {R1, R2};
      \path (R) edge[->, bend right=10] (R3);
    \end{tikzpicture}
    \caption{RL in an Unknown MDP}
    \label{fig:rl-unknown-mdp}
\end{figure}
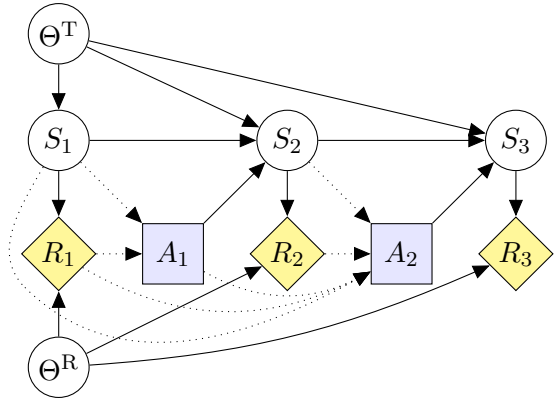

To model an \emph{unknown} MDP, we add additional parameters $\tf$ and $\rf$ to the transition and reward functions, which hold what is unknown about these functions.
Thus, for example, $R_i$ is now drawn from $R(S_i\,;\,\rf)$.
We extend the CID with additional variables for these unknown parameters, as shown in \cref{fig:rl-unknown-mdp}.
Now information links from previous rewards and actions are essential, as past states and rewards provide valuable information about the transition and reward function parameters.

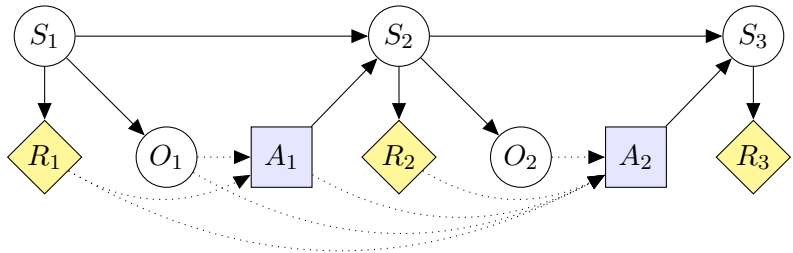
\begin{figure}[ht]
    \centering
    \begin{tikzpicture}[
        node distance=0.7cm,
        every node/.style={draw, circle, minimum size=0.8cm, inner sep=0.5mm}]
        \node (S1) {$S_1$};
        \node (R1) [below = of S1, utility] {$R_1$};
        \node (O1) [right = of R1] {$O_1$};
        \node (A1) [right = of O1, decision] {$A_1$};
        \node (S2) [right = 10em of S1] {$S_2$};
        \node (R2) [below = of S2, utility] {$R_2$};
        \node (O2) [right = of R2] {$O_2$};
        \node (A2) [right = of O2, decision] {$A_2$};
        \node (S3) [right = 10em of S2] {$S_3$};
        \node (R3) [below = of S3, utility] {$R_3$};
        
        \edge {S1} {S2,O1,R1};
        \edge {S2} {S3,O2,R2};
        \edge[information] {O1} {A1};
        \edge[information] {O2} {A2};
        \edge {S3} {R3};
        \edge {A1} {S2};
        \edge {A2} {S3};
        \path
          (R1) edge[->, information, bend right] (A1)
          (R1) edge[->, information, bend right] (A2)
          (O1) edge[->, information, bend right] (A2)
          (A1) edge[->, information, bend right] (A2)
          (R2) edge[->, information, bend right] (A2)
          ;
          
    \end{tikzpicture} 
    \caption{RL in a POMDP}
    \label{fig:rl-pomdp}
\end{figure}

To model a partially observed MDP, we add new variables $O_i$ representing the agent's observations
and remove the information links from the states, as shown in \cref{fig:rl-pomdp}.
Information links from past rewards and observations are included because they provide information about the unobserved states even when the transition and reward functions are known.
A POMDP with unknown parameters can be modeled explicitly with a similar transformation as from \cref{fig:rl-mdp} to \cref{fig:rl-unknown-mdp}, but it is simpler to use the same diagram as \cref{fig:rl-pomdp} and let the unobserved state variables $S_i$ hold the unobserved parameters $\tf$ and $\rf$.

\subsection{Current-RF Optimization}

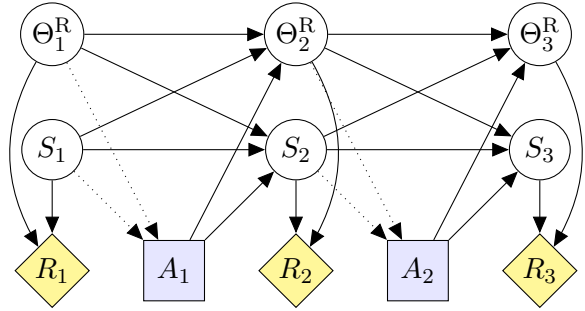
\begin{figure}[ht]
    \centering
    \begin{tikzpicture}[
        node distance=0.7cm,
        every node/.style={
            draw, circle, minimum size=0.8cm, inner sep=0.5mm}]
        \node (R1) [utility] {$R_1$};
        \node (S1) [above = of R1] {$S_1$};
        \node (A1) [right = of R1, decision] {$A_1$};
        \node (RF1) [above = of S1] {$\rf_1$};
        
        \node (R2) [right = of A1, utility] {$R_2$};
        \node (S2) [above = of R2] {$S_2$};
        \node (A2) [right = of R2, decision] {$A_2$};
        \node (RF2) [above = of S2] {$\rf_2$};
        
        \node (R3) [right = of A2, utility] {$R_3$};
        \node (S3) [above = of R3] {$S_3$};
        \node (RF3) [above = of S3] {$\rf_3$};
        
        \edge {S1} {R1};
        \edge {S2} {R2};
        \edge {S3} {R3};
        \edge {A1,S1,RF1} {S2};
        \edge {A2,S2,RF2} {S3};
        \edge {A1} {RF2};
        \edge {A2} {RF3};
        \edge[information] {S1} {A1};
        \edge[information] {S2} {A2};
        \edge[information] {RF1} {A1};
        \edge[information] {RF2} {A2};
        \edge {RF1} {RF2};
        \edge {RF2} {RF3};
        \edge {S1} {RF2};
        \edge {S2} {RF3};
        
        \path
        (RF1) edge[->, bend right] (R1)
        (RF2) edge[->, bend left] (R2)        
        (RF3) edge[->, bend left] (R3)
        ;
    \end{tikzpicture}
    \caption{RL in an MDP with a Modifiable Reward Function}
    \label{fig:rl-modifiable-reward}
\end{figure}

In the basic MDP from \cref{fig:rl-mdp}, the reward parameter $\rf$ is assumed to be unchanging.
In reality, this assumption may fail because the reward function is computed by some physical system that is a modifiable part of the state of the world.
\Cref{fig:rl-modifiable-reward} shows how
we can modify the CID to remove the assumption that the reward function is unchanging, by adding random variables $\rf_i$ representing the reward parameter at each time step.

\begin{figure}[ht]
    \centering
    \begin{tikzpicture}[
        node distance=0.7cm,
        every node/.style={
            draw, circle, minimum size=0.8cm, inner sep=0.5mm}]
        \node (R1) [utility] {$R_1$};
        \node (S1) [above = of R1] {$S_1$};
        \node (A1) [right = of R1, decision] {$A_1$};
        \node (RF1) [above = of S1] {$\rf_1$};
        
        \node (R2) [right = of A1, utility] {$R_2$};
        \node (S2) [above = of R2] {$S_2$};
        \node (A2) [right = of R2, decision] {$A_2$};
        \node (RF2) [above = of S2] {$\rf_2$};
        
        \node (R3) [right = of A2, utility] {$R_3$};
        \node (S3) [above = of R3] {$S_3$};
        \node (RF3) [above = of S3] {$\rf_3$};
        
        \edge {S1} {R1};
        \edge {S2} {R2};
        \edge {S3} {R3};
        \edge {A1,S1,RF1} {S2};
        \edge {A2,S2,RF2} {S3};
        \edge {A1} {RF2};
        \edge {A2} {RF3};
        \edge[information] {S1} {A1};
        \edge[information] {S2} {A2};
        \edge[information] {RF1} {A1};
        \edge[information] {RF2} {A2};
        \edge {RF1} {RF2};
        \edge {RF2} {RF3};
        \edge {S1} {RF2};
        \edge {S2} {RF3};
        
        \path
        (RF1) edge[->, bend right] (R1)
        ;
        \node (ah) [minimum size=0mm,node distance=2mm, below left = of R1, draw=none] {};
        \draw
        (RF1) edge[ in=145,out=-120] (ah.center)
        (ah.center) edge[->, out=-35,in=-150] (R2);
        \draw
        (RF1) edge[ in=145,out=-120] (ah.center)
        (ah.center) edge[->, out=-35,in=-150] (R3);
    \end{tikzpicture}
    \caption{Current-RF Optimization}
    \label{fig:sim-opt}
\end{figure}
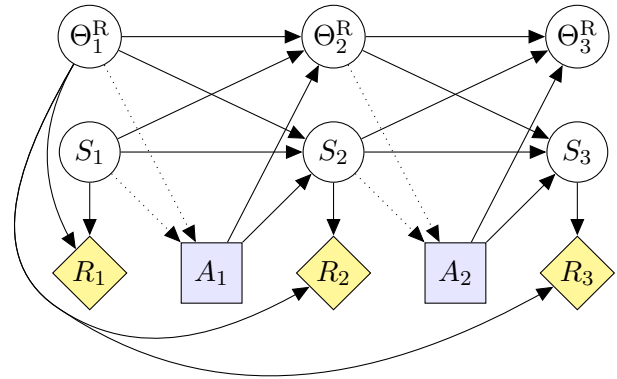

Now at each time step $i$, the agent receives reward $R_i = R(S_i\,;\,\rf_i)$.
This gives an incentive for the agent to obtain more reward by influencing the reward function rather than optimizing the state, sometimes called \emph{wireheading}.
An elegant solution to this problem is to use model-based agents
that simulate the state sequence likely to result from different policies, and evaluate those state sequences according to the current or initial reward function \citep{Everitt2018thesis,Orseau2011,Hibbard2012,Schmidhuber2007,Everitt2016sm}.
In contrast to RL in an unknown MDP, this is most easily implemented when the agent knows the reward function parameter.
The resulting CID is shown in \cref{fig:sim-opt}.


\subsection{Reward Modeling}

A key challenge when scaling RL to environments beyond board games or computer games is that it is hard to define good reward functions.
Reward Modeling~\citep{Leike2018alignment} is a safety framework in which the agent learns a reward model from human feedback while interacting with the environment.
The feedback could be in the form of preferences, demonstrations, real-valued rewards, or reward sketches.

We represent the feedback data using new variables $D_i$ which are used to train a reward model $M$.
Thus, the reward at timestep $i$ is given by $R_i=M(S_i\mid{D_1,\dots,D_{i-1}})$.
The feedback is produced based on past trajectories and unobserved human preferences $\hf$ by some data-generation process $G$.
Thus, the feedback data at timestep $i$ is given by 
$D_i=G(S_1,A_1,\dots,S_{i},A_{i}\,;\,\hf)$.

\begin{figure}[ht]
\begin{center}
  \begin{tikzpicture}[node distance=0.6cm]
      \tikzset{
        every node/.style={ draw, circle, minimum size=0.8cm, inner sep=0.5mm }
      }
      \node (R1) [utility] {$R_1$};
      \node (S1) [above = of R1] {$S_1$};
      \node (A1) [right = of R1, decision] {$A_1$};

      \node (R2) [right = of A1, utility] {$R_2$};
      \node (S2) [above = of R2] {$S_2$};
      \node (A2) [right = of R2, decision] {$A_2$};

      \node (R3) [right = of A2, utility] {$R_3$};
      \node (S3) [above = of R3] {$S_3$};
      
      \path
      (S1) edge[->] (R1)
      (S1) edge[->, information] (A1)
      (R1) edge[->, information] (A1)

      (S1) edge[->] (S2)
      (A1) edge[->] (S2)
      (R1) edge[->, information, bend right] (A2)

      (S2) edge[->] (R2)
      (S2) edge[->, information] (A2)
      (R2) edge[->, information] (A2)

      (S2) edge[->] (S3)
      (A2) edge[->] (S3)

      (S3) edge[->] (R3)
      ;
      \node (ah) [minimum size=0mm,node distance=2mm, below left = of R1, draw=none] {};
      \draw[information]
      (S1) edge[ in=145,out=-130] (ah.center)
      (ah.center) edge[->, out=-25,in=-150] (A2);
      
      \node (D1) [below = of A1] {$D_1$};
      \node (D2) [below = of A2] {$D_2$};
      \node (help) [below = of D1, draw=none] {};
      \node (Theta) at (R2 |- help) []  {$\hf$};
      
      \node (space) [minimum size=0mm, node distance=2mm, below left = 1em of D1, draw=none] {};
      \draw (S1) edge[in=155,out=-150] (space.center)
      (space.center) edge[->,out=-25,in=-150] (D2);
      
      \path
      (S1) edge[->]  (D1)
      (S2) edge[->] (D2)
      (D1) edge[->] (R2)
      (D2) edge[->] (R3)
      (D1) edge[->, out=5, in=220] (R3)
      (Theta) edge[->] (D1)
      (Theta) edge[->] (D2)
      ;
      \edge{A1}{D1,D2};
      \edge{A2}{D2};
    \end{tikzpicture}
\end{center}
\caption{Reward Modeling}\label{fig:rm}
\end{figure}
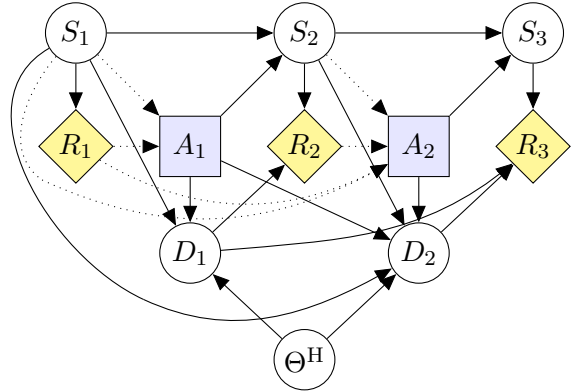


The CID in \cref{fig:rm} is a variation of the basic MDP diagram of \cref{fig:rl-mdp}, where variables $D_i$ and $\hf$ are added for the training of the reward model.
A parameter $\Theta^{\mathrm{T}}$ can be added to model an unknown transition function as in \cref{fig:rl-unknown-mdp}.

Reward modeling can also be done recursively, using previously trained agents to help with the training of more powerful agents \citep{Leike2018alignment}.
In this case, the previously trained agents can be considered to be part of the data generation process $G$, so are captured by the $D_i$ variables.


\subsection{CIRL}

\begin{figure}[ht]
\begin{center}
    \begin{tikzpicture}[
        node distance=0.7cm,
        every node/.style={
            draw, circle, minimum size=0.8cm, inner sep=0.5mm}]
        \node (R1) [utility] {$R_1$};
        \node (S1) [above = of R1] {$S_1$};
        \node (A1) [right = of R1, decision, player1, fill=blue!10] {$A_1$};
        \node (R2) [right = of A1, utility] {$R_2$};
        \node (S2) [above = of R2] {$S_2$};
        \node (A2) [right = of R2, decision, player1, fill=blue!10] {$A_2$};
        \node (R3) [right = of A2, utility] {$R_3$};
        \node (S3) [above = of R3] {$S_3$};
        
        \edge {S1} {R1};
        \edge {S2} {R2};
        \edge {S3} {R3};
        \edge {A1,S1} {S2};
        \edge {A2,S2} {S3};
        \edge[information] {S1} {A1};
        \edge[information] {S2} {A2};
        
        \node (A1h) [below = of A1, decision, player2] {$A^\mathrm{H}_1$};
        \node (A2h) [below = of A2, decision, player2] {$A^\mathrm{H}_2$};
        
        \node (help) [below = of A1h, draw=none] {};
        
        \node (theta) at (R2 |- help) [] {$\hf$};
        
        \node (space) [minimum size=0mm, node distance=2mm, below left = 1em of A1h, draw=none] {};
        \draw (S1) edge[information,in=155,out=-150] (space.center)
        (space.center) edge[information,->,out=-25,in=-150] (A2h);
          
        \path 
        (A1h) edge[->] (S2)
        (A1h) edge[->, information] (A1)
        (A1h) edge[->, information, bend right=7] (A2)        
        (A2h) edge[->] (S3)
        (A2h) edge[->, information] (A2)
        
        (A1) edge[->, information, bend right] (A2)
        
        (S1) edge[->, information] (A1h)
        (S2) edge[->, information] (A2h)
        (A1) edge[->, information] (A2h)
        (A1h) edge[->, information] (A2h)
        (theta) edge[->, information] (A1h)
        (theta) edge[->, information] (A2h)        
        
        (theta) edge[->, out=180, in=270] (R1)
        (theta) edge[->] (R2)
        (theta) edge[->, out=0, in=-90] (R3)
        ;
        
      \node (ah) [minimum size=0mm,node distance=2mm, below left = of R1, draw=none] {};
      \draw[information]
      (S1) edge[ in=145,out=-130] (ah.center)
      (ah.center) edge[->, out=-25,in=-150] (A2);
    \end{tikzpicture}
\end{center}
\caption{CIRL. The human's actions are in green and the agent's in blue; both try to optimize the yellow rewards.}\label{fig:cirl}
\end{figure}
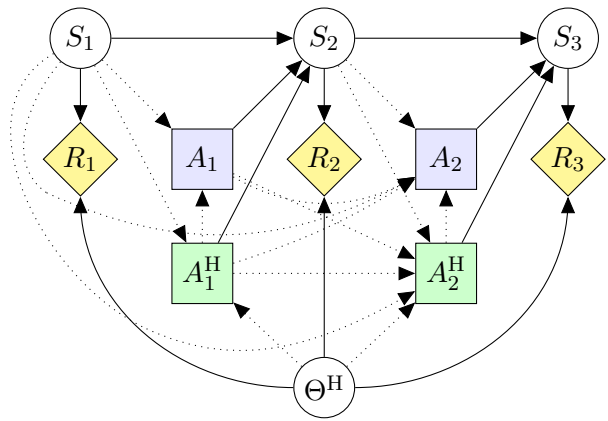

Another way for agents to learn the reward function while interacting with the environment is Cooperative Inverse Reinforcement Learning (CIRL) \citep{Hadfield-Menell2016cirl}. 
Here the agent and the human inhabit a joint environment.
The human and the agent jointly optimize the sum of rewards, but only the human knows what the rewards are.
The agent has to infer the rewards by looking at the human's actions.
We represent this by adding two kinds of variable to the basic MDP diagram from \cref{fig:rl-mdp}: the human's action $A_i^\mathrm{H}$ at timestep $i$, and a parameter $\hf$ controlling the reward function (known to the human but unknown to the agent).
The resulting CID is shown in \cref{fig:cirl}.


CIRL is closely related to reward modeling (\cref{fig:rm}), if we read the human actions $A_i^\mathrm{H}$ as the human data $D_i$.
However, the edges are not the same: in CIRL the human's actions are observed by the agent and affect the next state, whereas in reward modeling the feedback data affects the rewards.
Likewise, the CIRL diagram has edges directly from $\hf$ to $R_i$, since the rewards are produced by the true reward function rather than a reward model.
In reward modeling, the path from from $\hf$ to $R_i$ is mediated by $D_i$.
Another difference is that in CIRL, the rewards are never shown to the agent, which may remain uncertain about how much reward it generated even at the very end of the episode.

In principle, information links from past rewards to the human's action nodes $A_i^\mathrm{H}$ could be added to \cref{fig:cirl}, but they are unnecessary since the human knows the reward parameter $\hf$.
Unlike the current-RF optimization setup (\cref{fig:sim-opt}), the reward parameter is not observed by the agent.


\section{Question-Answering Systems}
\label{sec:qa-systems}

Outside of the MDP-based frameworks, a common way to use machine learning systems is to train them to answer questions. Supervised learning fits in this question-answering framework.
For example, in image recognition, the question is an image, and the answer is a label for the image such as a description of the main object depicted.
Other applications that fit this paradigm include machine translation, speech synthesis, and calculation of mortgage default probabilities.
In this section we consider safety frameworks that fit in the question-answering (QA system) paradigm.

A literal interpretation of a QA-system is a system to which the user submits a query about the future and receives an answer in plain text.
Such a system can help a trader predict stock prices, help a doctor make a diagnosis, or help a politician choose the right policy against climate change. QA-systems have some safety advantages over the MDP-based systems described in \cref{sec:mdp-based}, as they do not directly affect the world and do not engage in long-term planning.

\subsection{Supervised Learning}

In supervised learning, the question corresponds to the input and the state corresponds to the output label. 
Since the training labels are generated independently, they are not affected by the agent's actions.
Once the system is deployed, it keeps acting \emph{as if} its answer does not affect the label at deployment, whether or not it does. 
Theorem proving is another application where this structure holds: a proposed proof for a theorem can be automatically checked.  
The setup is shown \cref{fig:tool-ai}. 
It is sometimes called Tool AI \citep{bostrom2014superintelligence,Gwern2016}.
So called Act-based agents share the same causal structure \citep{Christiano2015a}.


%

\ifijcai\else
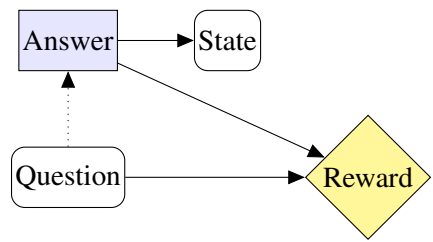
\begin{figure}[ht]
\centering
    \begin{tikzpicture}[
    node distance=1cm,
    every node/.style={
      draw, rectangle, rounded corners=5, minimum size=0.8cm, inner sep=0.5mm, align=center}]
      
      \node (Q) {Question};
      
      \node (A) [above = of Q, decision] {Answer};
      \node (S) [right = of A] {State};
      
      \node (rs) [right = of S, draw=none] {};
      
      \node (R) at (rs |- Q) [utility] {Reward};
      
      \edge[->, information] {Q} {A};
      \edge[->] {A} {S};
      \edge[->] {Q, A} {R};
   \end{tikzpicture}
   \caption{Supervised Learning}
   \label{fig:tool-ai}
\end{figure}
\fi

\subsection{Self-Fulfilling Prophecies}

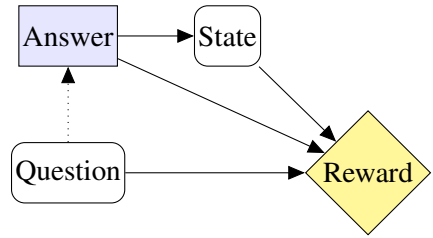
\begin{figure}[ht]
\centering
    \begin{tikzpicture}[
    node distance=1cm,
    every node/.style={
      draw, rectangle, rounded corners=5, minimum size=0.8cm, inner sep=0.5mm, align=center}]
      
      \node (Q) {Question};
      
      \node (A) [above = of Q, decision] {Answer};
      \node (S) [right = of A] {State};
      
      \node (rs) [right = of S, draw=none] {};
      
      \node (R) at (rs |- Q) [utility] {Reward};
      
      \edge[->, information] {Q} {A};
      \edge[->] {A} {S};
      \edge[->] {Q, A} {R};
      \path (S) edge[->] (R);
   \end{tikzpicture}
   \caption{Supervised Learning with Self-Fulfilling Prophecies}
   \label{fig:oracle}
\end{figure}

The assumption that the labels are generated independently of the agent's answer sometimes fails to hold.
For example, the label for an online stock price prediction system could be produced after trades have been made based on its prediction.
In this case, the QA-system has an incentive to make \emph{self-fulfilling prophecies}.
For example, it may predict that the stock will have zero value in a week.
If sufficiently trusted, this prediction may lead the company behind the stock to quickly go bankrupt. 
Since the answer turned out to be accurate, the QA-system would get full reward.
This problematic incentive is represented in the diagram in \cref{fig:oracle}, where we can see that the QA-system has both incentive and ability to affect the world state with its answer \citep{Everitt2019incentives1}.

\ifijcai
\begin{figure}[t]
\centering
    \begin{tikzpicture}[
    node distance=1cm,
    every node/.style={
      draw, rectangle, rounded corners=5, minimum size=0.8cm, inner sep=0.5mm, align=center}]
      
      \node (Q) {Question};
      
      \node (A) [above = of Q, decision] {Answer};
      \node (S) [right = of A] {State};
      
      \node (rs) [right = of S, draw=none] {};
      
      \node (R) at (rs |- Q) [utility] {Reward};
      
      \edge[->, information] {Q} {A};
      \edge[->] {A} {S};
      \edge[->] {Q, A} {R};
   \end{tikzpicture}
   \caption{Supervised Learning}
   \label{fig:tool-ai}
\end{figure}
\fi

\ifijcai
\begin{figure}[ht]
\centering
\begin{tikzpicture}[
  node distance=1cm,
  every node/.style={
    draw, rectangle, rounded corners=5, minimum size=0.8cm, inner sep=0.5mm, align=center}]
  
  \node (Q) {Question};
  
  \node (A) [above = of Q, decision] {Answer};
  \node (S) [right = of A] {State};
  
  \node (rs) [right = of S, draw=none] {};
  
  \node (R) at (rs |- Q) [utility] {Reward};
  
  \edge[->, information] {Q} {A};
  \edge[->] {A} {S};
  
  \node [below = of Q, decision, rectangle, rounded corners=5,counterfactual] (cA) {Answer \\ hidden};
  \node at (S |- cA) (cS) [counterfactual] {Counter-\\factual\\ state};
  
 \edge[->] {Q} {cA};
 \edge[->] {cA} {cS};
 \edge[->] {Q, A, cS} {R};
\end{tikzpicture}
\caption{Counterfactual Oracle. Dashed nodes represent counterfactual variables.}
\label{fig:counterfactual-oracle}
\end{figure}
\fi

\subsection{Counterfactual Oracles}

It is possible to fix the incentive for making self-fulfilling prophecies while retaining the possibility to ask questions where the correctness of the answer depends on the resulting state.
\emph{Counterfactual oracles} optimize reward in the counterfactual world where no one reads the answer \citep{Armstrong2017oracles}.
This solution can be represented with a \emph{twin network} \citep{Balke1994} influence diagram, as shown in \cref{fig:counterfactual-oracle}.
Here, we can see that the QA-system's incentive to influence the (actual) world state has vanished, since the actual world state does not influence the QA-system's reward; thereby the incentive to make self-fulfilling prophecies also vanishes.
We expect this type of solution to be applicable to incentive problems in many other contexts as well.
A concrete training procedure for counterfactual oracles can also be represented with influence diagrams \citep[Section~4.4]{Everitt2019incentives1}.

\ifijcai\else
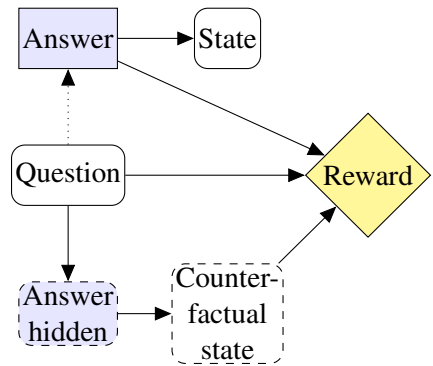
\begin{figure}[ht]
\centering
\begin{tikzpicture}[
  node distance=1cm,
  every node/.style={
    draw, rectangle, rounded corners=5, minimum size=0.8cm, inner sep=0.5mm, align=center}]
  
  \node (Q) {Question};
  
  \node (A) [above = of Q, decision] {Answer};
  \node (S) [right = of A] {State};
  
  \node (rs) [right = of S, draw=none] {};
  
  \node (R) at (rs |- Q) [utility] {Reward};
  
  \edge[->, information] {Q} {A};
  \edge[->] {A} {S};
  
  \node [below = of Q, decision, rectangle, rounded corners=5,counterfactual] (cA) {Answer \\ hidden};
  \node at (S |- cA) (cS) [counterfactual] {Counter-\\factual\\ state};
  
 \edge[->] {Q} {cA};
 \edge[->] {cA} {cS};
 \edge[->] {Q, A, cS} {R};
\end{tikzpicture}
\caption{Counterfactual Oracle. Dashed nodes represent counterfactual variables.}
\label{fig:counterfactual-oracle}
\end{figure}
\fi



\subsection{Debate}

The QA-systems discussed so far all require that it is possible to check whether the agent's answer was correct or not.
However, this can be difficult in some important applications.
For example, we may wish to ask the system about the best way to solve climate change if we want a healthy climate on earth 100 years from now.
In principle we could follow the system's suggestion for 100 years, and see whether the answer was right.
However, in practice waiting 100 years before rewarding the system or not is slow (too slow for training the system, perhaps), and the cost of following bad advice would be substantial.

To fix this, \citet{Irving2018debate} suggest pitting two QA-systems against each other in a debate about the best course of action.
The systems both make their own proposals, and can subsequently make arguments about why their own suggestion is better than their opponent's.
The system who manages to convince the user gets rewarded; the other system does not.
While there is no guarantee that the winning answer is correct, the setup provides the user with a powerful way to poke holes in any suggested answer, and reward can be dispensed without waiting to see the actual result.

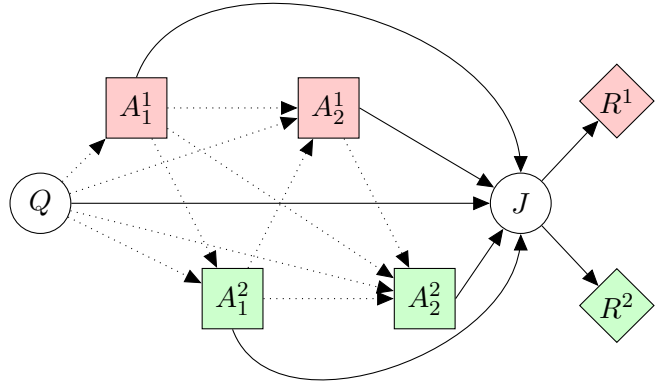
\begin{figure}[ht]
    \centering
    \begin{tikzpicture}[
        node distance=0.45cm,
        every node/.style={
            draw, circle, minimum size=0.8cm, inner sep=0.5mm}]

        \node (Q) [] {$Q$};
        \node (space1) [right = of Q, draw=none] {};
        \node (space2) [right = of space1, draw=none] {};
        \node (space3) [right = of space2, draw=none] {};
        \node (space4) [right = of space3, draw=none] {};
        
        \node (A11) [decision, above = of space1, player1] {$A_1^1$};
        \node (A21) [decision, below = of space2, player2] {$A_1^2$};
        \node (A12) [decision, above = of space3, player1] {$A_2^1$};
        \node (A22) [decision, below = of space4, player2] {$A_2^2$};
        
        \node (J) [right = of space4] {$J$};
        \node (space5) [right = of J, draw=none] {};

        \node (R1) [utility, above = of space5, player1] {$R^1$};
        \node (R2) [utility, below = of space5, player2] {$R^2$};
        
        \edge [information] {Q} {A11, A12, A21, A22};
        \edge [information] {A11} {A12, A21, A22};
        \edge [information] {A21} {A12, A22};
        \edge [information] {A12} {A22};
        \edge {A12.east, A22.east} {J};
        \path (A11.north) edge[out=70,in=90, ->] (J);
        \path (A21.south) edge[in = -90, out=-70, ->] (J);
        \edge {Q} {J};
        \edge {J} {R1, R2};
    \end{tikzpicture}
    \caption{Debate. Red decision and utility nodes belong to QA system 1, and green ones belong to QA system 2.}
    \label{fig:debate}
\end{figure}

The debate setup is shown in \cref{fig:debate}, where $Q$ represents the user's question, $A_i^k$ the $i$th statement by system $k\in \{1,2\}$.
Reward $R^k$ is given to system $k$ depending on the user's judgment $J$.


\subsection{Supervised IDA}

Iterated distillation and amplification (IDA) \citep{Christiano2018ida} is another suggestion that can be used for training QA-systems to correctly answer questions where it is hard for an unaided user to directly determine their correctness.
Given an original question $Q$ that is hard to answer correctly,
less powerful systems $X_k$ are asked to answer a set of simpler questions $Q_i$. 
By combining the answers $A_i$ to the simpler questions $Q_i$, the user can guess the answer $\hat A$ to $Q$.
A more powerful system $X_{k+1}$ is trained to answer $Q$, with $\hat A$ used as an approximation of the correct answer to $Q$.

Once the more powerful system $X_{k+1}$ has been trained, the process can be repeated. 
Now an even more powerful QA-system $X_{k+2}$ can be trained, by using $X_{k+1}$ to answer simpler questions to provide approximate answers for training $X_{k+2}$.
Systems may also be trained to find good subquestions, and for aggregating answers to subquestions into answer approximations. In addition to supervised learning, IDA can also be applied to reinforcement learning.

\begin{figure}[ht]
\centering
\begin{tikzpicture}[
    node distance=0.9cm and 0.7cm,
    every node/.style={
        draw,circle, minimum size=0.8cm, inner sep=0.5mm}]
    \node (Q) [] {$Q$};
    \node (Q1) [below right = of Q] {$Q_1$};
    \node (A1) [right = of Q1] {$A_1$};
    \node (Q2) [below = of Q1] {$Q_2$};
    \node (A2) [right = of Q2] {$A_2$};

    \node (A) [above right = of A1] {$\hat A$};
    \node (X) [above = of Q, decision] {$A$};
    \node (R) at (A|-X) [utility] {$R$};
    
    \edge {Q} {Q1, Q2};
    \edge {Q1} {A1};
    \edge {Q2} {A2};
    
    \path 
    (Q2) edge[->, bend right=20] (A)
    (A2) edge[->, bend right=10] (A)
    ;
    
    \path
    (Q) edge[->, information] (X)
    (Q) edge[->] (Q1)
    (Q) edge[->] (A)
    (Q1) edge[->] (A1)
    (Q1) edge[->] (A)
    (A1) edge[->] (A)
    (A) edge[->] (R)
    (X) edge[->] (R)
    ;
\end{tikzpicture}
\caption{Supervised IDA}
\end{figure}
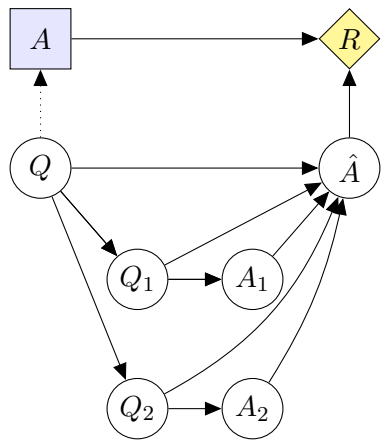

\ifijcai
\begin{figure}[t]
\centering
\begin{tikzpicture}[
    node distance=0.9cm and 0.7cm,
    every node/.style={
        draw,circle, minimum size=0.8cm, inner sep=0.5mm}]
        
    \node (I) {Input};
    \node (A1) [right = of I, decision, player1] {$A^1$};
    \node (R1) [above = of A1, utility, player1] {$R^1$};
    
    \node (R2) [below = of A1, utility, player2] {$R^2$};
    \node (A2) [below = of R2, decision, player2] {$A^2$};
    
    \node (R3) [right = of R1, utility, player3] {$R^3$};
    \node (A3) [right = of A1, decision, player3] {$A^3$};
    
    \node (R4) [right = of R2, utility, player4] {$R^4$};
    \node (A4) [right = of A2, decision, player4] {$A^4$};
    
    \node (O) [right = of A3] {Output};
    
    \edge {I} {R1, R2};
    \edge {A1} {R1, R3};
    \edge {A2} {R2, R4};
    \edge {A3} {R3, O};
    \edge {A4} {R4, O};
    
    \edge[information] {I} {A1, A2};
    \edge[information] {A1} {A3};
    \edge[information] {A2} {A4};
      
\end{tikzpicture}
\caption{Comprehensive AI Services. The decision and utility nodes of each service is shown with a different colour and variable superscript.}
\label{fig:cais}
\end{figure}
\fi

\subsection{Comprehensive AI Services}

\citet{Drexler2019} argues that the main safety concern from artificial intelligence does not come from a single agent, but rather from big collections of AI services.
For example, one service may provide a world model, another provide planning ability, a third decision making, and so on.
As an aggregate, these services can be very competent, even though each service only has access to a limited amount of resources and only optimizes a short-term goal.

A simple model of Drexler's \emph{comprehensive AI services} (CAIS) is shown in \cref{fig:cais}, where the output of one service can be used as input to another.
The general CAIS framework also allows services that develop and train other services.

\ifijcai\else
\begin{figure}[ht]
\centering
\begin{tikzpicture}[
    node distance=0.9cm and 0.7cm,
    every node/.style={
        draw,circle, minimum size=0.8cm, inner sep=0.5mm}]
        
    \node (I) {Input};
    \node (A1) [right = of I, decision, player1] {$A^1$};
    \node (R1) [above = of A1, utility, player1] {$R^1$};
    
    \node (R2) [below = of A1, utility, player2] {$R^2$};
    \node (A2) [below = of R2, decision, player2] {$A^2$};
    
    \node (R3) [right = of R1, utility, player3] {$R^3$};
    \node (A3) [right = of A1, decision, player3] {$A^3$};
    
    \node (R4) [right = of R2, utility, player4] {$R^4$};
    \node (A4) [right = of A2, decision, player4] {$A^4$};
    
    \node (O) [right = of A3] {Output};
    
    \edge {I} {R1, R2};
    \edge {A1} {R1, R3};
    \edge {A2} {R2, R4};
    \edge {A3} {R3, O};
    \edge {A4} {R4, O};
    
    \edge[information] {I} {A1, A2};
    \edge[information] {A1} {A3};
    \edge[information] {A2} {A4};
      
\end{tikzpicture}
\caption{Comprehensive AI Services. The decision and utility nodes of each service is shown with a different colour and variable superscript.}
\label{fig:cais}
\end{figure}
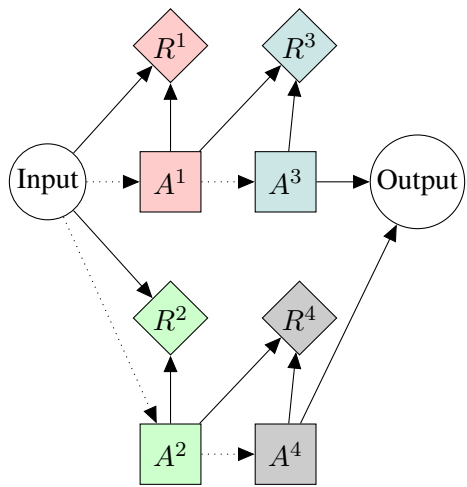
\fi

\section{Discussion}
\label{sec:discussion}

Causal influence diagrams of frameworks provide a useful perspective for understanding AGI safety problems and solutions, but there are some subtleties to be explored in their use and interpretation.
In this section we discuss some of these issues.
First, we look at CIDs' focus on agents and potential problems with this view.
Then we discuss the use and interpretation of CIDs, and the flexibility and design choices one has when modeling frameworks.

\subsection{Intentional Stance}

Causal influence diagram representations are inherently agent-centric.
Their key assumption is that one or more intelligent systems are making decisions in a causal environment in order to maximize the sum of some real-valued variables.
This is an important perspective for AGI safety,
as increasingly powerful optimization processes are a core component of most AI and machine learning methods,
and many AGI safety problems are consequences of such optimization processes becoming too powerful in some sense.
The agent perspective is especially relevant for RL, perhaps the most promising method for constructing AGI.

A CID models an agent that ``knows'' the framework it is facing.
Either it has been designed for that framework (Q-learning has been designed to interact with an MDP), or it has been pre-trained for the framework (e.g.\ by meta-learning~\citep{Wang2017meta}).
The agent's utility is a kind of fiction, as is easy to see when all learning is offline: at deployment, the policy parameters are frozen and the agent executes a fixed pre-trained policy without receiving reward signals.
However, the agent still acts \emph{as if} it optimizes the reward function it had during training, which means that the incentives arising from this fictional reward can still be used to predict actual agent behaviour.

We adopt the \emph{intentional stance}~\citep{Dennett1987,Orseau2018} towards the agent, which means that it is not important whether the agent's utility has an obvious physical correlate (e.g., the value of some machine register, which may apply to an RL agent learning an MDP online).
What counts is that treating the system as an agent optimizing the reward is a good model for predicting its behaviour.
We take this stance because present-day systems already admit it, and because we think it will be increasingly admissible for more advanced artificial agents.

Of course, the intentional stance is not always the most useful perspective for AGI safety questions.
For example, concerns about bounded rationality or adversarial inputs to an imperfect system do not show up prominently when we analyze the incentives of a rational agent in some framework.
It can also be tricky to model the evolution of a system that is initially agent-like, but through self-modifications or accidental breakage turns into something less goal-directed.
Finally, some safety frameworks include insights that are not fully captured by the agent-centric view: the CAIS framework can be seen as a collection of agents, but the overall system does not naturally fit into this view.

\subsection{Modeling Choices}

Just like a causal graph \citep{Pearl2009}, a CID implies many causal relationships and conditional independencies between nodes.
A model is only accurate if it gets those relationships right.
Generally, there should be a directed path between a node $X$ and a node $Y$ if and only if an exogenous intervention changing the outcome of $X$ could affect the outcome of $Y$.
Similarly, if two nodes $X$ and $Y$ are conditionally dependent when conditioning on a third set $\mathbf{Z}$ of nodes, then $X$ and $Y$ should not be d-separated in the graph when conditioning on $\mathbf{Z}$ \citep{Pearl2009}.

These requirements already put heavy constraints on what can be seen as an accurate CID of a given framework.
However, some choices are usually still left to the modeler.
Most notably, freedom remains in whether to aggregate or split random variables.
For example, in the MDP in \cref{fig:rl-mdp}, a single node represents the state, while the MDP with a modifiable reward function in \cref{fig:rl-modifiable-reward} can be seen as an MDP where state is represented by two nodes $S_i$ and $\rf_i$.
The MDP diagram is more \emph{coarse-grained} and the MDP with a modifiable reward function is more \emph{fine-grained}.
In general, there is a trade-off between representing many details and highlighting the big picture~\citep{Hoel2013causal}.

Another choice is which random variables should be represented at all.
For example, in reward modeling in \cref{fig:rm}, an extra node could have been added for the reward model itself.
However, we left this out to keep the diagram simple.
From a probability theory perspective, we can think of the variable representing the reward model having been \emph{marginalized out}.
It is not always possible to marginalize out nodes.
For example, it is not possible to marginalize out $\hf$ in reward modeling (\cref{fig:rm}), because $\hf$ is necessary for there to be a non-causal correlation between the feedback data nodes.

Two safety frameworks, reward modeling and IDA, include a recursive aspect that is intended to help them scale to settings where feedback data is not easily produced by unaided humans.
When modeling these frameworks, we could take the intentional stance towards previous iterations of the recursive setup, resulting in a multi-agent CID.
We found that this seemed to make the diagram complicated without adding much insight, since the execution of recursively trained agents can be considered an implementation detail.
Still, this shows that there can be choice about which aspects of a framework to model using the intentional stance.

%
%

\section{Conclusion}

In this paper, we have modeled many of the most prominent AGI safety frameworks with causal influence diagrams.
The unified representation allows us to pinpoint key similarities and differences.
For example, a key difference between CIRL and reward modeling is the causal relationship between the agent's reward and the user's preferences.
In CIRL, the reward follows directly from the user's preferences, whereas in reward modeling the path from user preferences to rewards is mediated by the feedback data.

Many of the important aspects for AGI safety are about causal relationships, and agents' decisions to influence their environment according to the objectives they are given.
This is exactly what causal influence diagrams describe, and what makes causal influence diagrams well suited to representing and comparing different safety frameworks.
For example, a well-known problem with prediction systems is that they incentivize self-fulfilling prophecies.
This is neatly represented with a causal path from the system's answer to the world state to the system's reward.
We could also see how various variants avoided the problem of self-fulfilling prophecies.
Both Tool AI and Counterfactual Oracles break the problematic causal path.
We expect future studies of AGI safety frameworks with causal influence diagrams to reveal further problems and suggest possible solutions.

\bibliography{references}{}
\bibliographystyle{named}

\end{document}